%% file: ms.tex
\pdfoutput=1

\documentclass[11pt]{article}

\usepackage{EACL2023}

\usepackage{times}
\usepackage{latexsym}

\usepackage[T1]{fontenc}

\usepackage[utf8]{inputenc}

\usepackage{microtype}

\usepackage{inconsolata}

\usepackage{graphicx}
\usepackage{amsmath}
\usepackage{amssymb}
\usepackage{bbding}
\usepackage{subcaption, booktabs} 
\usepackage{color, colortbl}
\usepackage{soul}
\usepackage{enumitem}
\usepackage{makecell}
\usepackage{multirow}
\usepackage{multicol}
\usepackage{url}
\usepackage{CJKutf8}
\usepackage{hyperref}
\usepackage[ruled]{algorithm2e}
\usepackage{hyperref}
\renewcommand{\footnotesize}{\fontsize{9pt}{11pt}\selectfont}
\usepackage{footmisc}

\usepackage{xcolor}
\definecolor{dark-green}{RGB}{84, 130, 53}
\definecolor{dark-red}{RGB}{192, 0, 0}
\definecolor{LimeGreen}{RGB}{50, 205, 50}

\title{Friend-training: Learning from Models of Different but Related Tasks}

\author{First Author \\
  Affiliation / Address line 1 \\
  Affiliation / Address line 2 \\
  Affiliation / Address line 3 \\
  \texttt{email@domain} \\\And
  Second Author \\
  Affiliation / Address line 1 \\
  Affiliation / Address line 2 \\
  Affiliation / Address line 3 \\
  \texttt{email@domain} \\}

\newcommand{\suda}{$^{\dagger}$}
\newcommand{\tencent}{$^{\diamond}$}

\author{Mian Zhang\suda\Thanks{~Work done when interning at Tencent AI Lab.},
        Lifeng Jin\tencent,
        Linfeng Song\tencent,
        Haitao Mi\tencent,
        Xiabing Zhou\suda \and
        Dong Yu\tencent \\
        \suda{}Soochow University, Suzhou, China\\
        \texttt{mzhang2@stu.suda.edu.cn}, \texttt{zhouxiabing@suda.edu.cn} \\
        \tencent{}Tencent AI Lab, Bellevue, WA, USA \\
        \texttt{\{lifengjin,lfsong,haitaomi,dyu\}@tencent.com}
}

\begin{document}
\maketitle
\begin{abstract}
Current self-training methods such as standard self-training, co-training, tri-training, and others often focus on improving model performance on a single task, utilizing differences in input features, model architectures, and training processes. However, many tasks in natural language processing are about different but related aspects of language, and models trained for one task can be great teachers for other related tasks. In this work, we propose \textbf{friend-training}, a \textit{cross-task} self-training framework, where models trained to do different tasks are used in an iterative training, pseudo-labeling, and retraining process to help each other for better selection of pseudo-labels. With two dialogue understanding tasks, conversational semantic role labeling and dialogue rewriting, chosen for a case study, we show that the models trained with the friend-training framework achieve the best performance compared to strong baselines.

\end{abstract}

\begin{CJK*}{UTF8}{gkai}
\input{introduction}
\input{related_work}
\input{framework}
\input{experiment}
\input{conclusion}
\input{limitation}

\end{CJK*}

\bibliography{emnlp2022}
\bibliographystyle{acl_natbib}

\appendix

\begin{CJK*}{UTF8}{gkai}
\input{appendix}
\end{CJK*}

\end{document}

%% file: introduction.tex
\section{Introduction}
Many different machine learning algorithms, such as self-supervised learning~\cite{Mikolov2013-mr,devlin2018bert,liu2021self}, semi-supervised learning~\cite{yang2021survey} and weakly supervised learning~\cite{zhou2018brief},  aim at using unlabeled data to boost performance.
They have been of even greater interest recently given the amount of unlabeled data available. Self-training~\cite{scudder1965probability} is one semi-supervised learning mechanism aiming to improve model performance through pseudo-labeling and has been successfully applied to computer vision~\cite{lee2013pseudo,chen2021semi}, natural language processing~\cite{dong2011ensemble,bhat2021self} and other fields~\cite{wang2019attributed,kahn2020self}.

\begin{figure}[!t]
\centering
\includegraphics[width=\columnwidth]{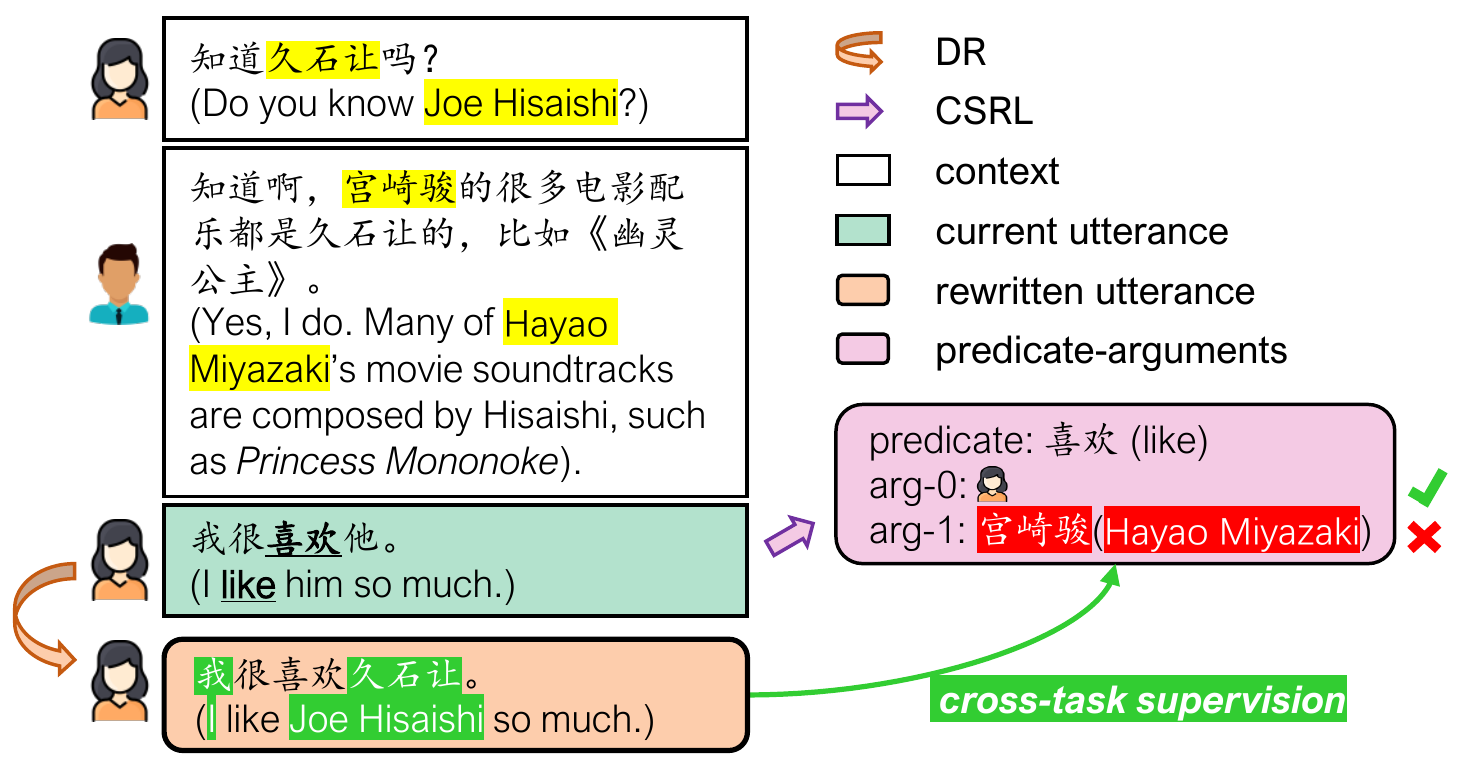}
\caption{An example of cross-task supervision between a CSRL parser and a DR system in a dialogue. 久石让(\colorbox{LimeGreen}{\textcolor{white}{Joe Hisaishi}}) from the rewritten utterance provides cross-task supervision to 宫崎骏(\colorbox{red}{\textcolor{white}{Hayao Miyazaki}}), the predicted \texttt{arg-1} of 喜欢(like) from the CSRL parser, while 我(\colorbox{LimeGreen}{\textcolor{white}{I}}) to the predicted \texttt{arg-0}.}\label{fig:intro}
\end{figure}

The main challenge of self-training is how to select high-quality pseudo-labels. Current self-training algorithms mainly focus on a single task when assessing the quality of pseudo-labels and suffer from gradual drifts of noisy instances~\cite{zhang2021understanding}. This work is motivated by the observation that learning targets of tasks represent different properties of the inputs, and some properties are shared across the tasks which can be used as supervision from one task to another. Such properties include certain span boundaries in dependency and constituency parsing, and some emotion polarities in sentiment analysis and emotion detection.
Two dialogue understanding tasks, conversational semantic role labeling (CSRL) and dialogue rewriting (DR), are also such a pair, with shared properties such as coreference and zero-pronoun resolution. As shown in Figure~\ref{fig:intro}, the rewritten utterance can be used to generate cross-task supervision to the arguments of predicate 喜欢 (like). We leverage the cross-task supervision from \textit{friend tasks} -- different but related tasks -- as a great criterion for assessing the quality of pseudo-labels.

In this work, we propose \textbf{friend-training}, the first \textit{cross-task} self-training framework. Compared to single-task self-training, friend-training \textbf{exploits supervision from friend tasks for better selection of pseudo-labels}. To this end, two novel modules are proposed: (1) a \textit{translation matcher}, which maps the pseudo-labels of different tasks for one instance into the same space and computes a \textit{matching score} representing the \textbf{cross-task matching degree of pseudo-labels from different tasks}; (2) an \textit{augmented (instance) selector}, which leverages \textbf{both} the confidence of pseudo-labels from task-specific models and the matching score to select instances with pseudo-labels of high quality as new training data. We choose CSRL and DR as friend tasks to conduct a case study for friend-training, and specify the translation matcher and augmented selector for friend-training between these tasks. Experimental results of domain generalization and few-shot learning show friend-training surpasses both classical and state-of-the-art semi-supervised learning algorithms by a large margin.
To summarize, contributions from this work include:
\begin{itemize}[itemsep=-1pt,topsep=6pt,leftmargin=10pt]
    \item We propose friend-training, the first cross-task self-training framework which exploits supervision from friend tasks for better selection of pseudo-labels in the iterative training process.
    \item We provide specific modeling of friend-training between CSRL and DR, with a novel translation matcher and a novel augmented selector. 
    \item Extensive experiments with CSRL and DR demonstrate the effectiveness of friend-training, outperforming several strong baselines.
\end{itemize}

%% file: related_work.tex
\section{Related Work}
\noindent\textbf{Self-training} Self-training~\cite{scudder1965probability,Angluin1988-yw,Abney2002-qo,lee2013pseudo} is a classical semi-supervised learning framework~\cite{chapelle2009semi} which has been widely explored in recent years. The general idea of self-training is to adopt a trained model to pseudo-label easily acquired unlabeled data and use them to augment the training data to retrain the model iteratively. This paradigm shows promising effectiveness in a variety of tasks: including text classification~\cite{mukherjee2020uncertainty,wang2020adaptive}, image classification~\cite{xie2020self,zoph2020rethinking}, machine translation~\cite{he2019revisiting} and model distillation~\cite{mukherjee2020xtremedistil}. Co-training~\cite{blum1998combining} and tri-training~\cite{zhou2005tri} are similar iterative training frameworks to self-training but with a different number of models or considering different views of the training data, both of which see wide adoption in NLP \cite{Mihalcea2004-ks,McClosky2006-mp,Wan2009-jq,Li2014-ie,Caragea2015-fm,Lee2021-uj,Wagner2021-qo}. These frameworks aim at improving performance with multiple models trained on one task, without directly leveraging the benefit of supervision from related tasks.\\
\noindent\textbf{Multi-task Learning} Multi-task learning~\cite{caruana1997multitask,yang2021survey} seeks to improve the learning performance of one task with the help of other related tasks, among which two lines of work are related to ours: (1) semi-supervised multi-task learning~\cite{liu2007semi,li2009active} combines semi-supervised learning and multi-task learning. \citet{liu2007semi} exploited unlabeled data by random walk and used a task clustering method for multi-task learning. \citet{li2009active} integrated active learning~\cite{mackay1992information} with the model in \citet{liu2007semi} to retrieve data that are most informative for labeling. Although these works tried to utilize unlabeled data to enhance multi-task learning, our work differs from them in incorporating supervised signals among tasks to select high-quality pseudo-labels for updating models, which is an iterative training process without additional human annotation. (2) Task grouping~\cite{kumar2012learning,standley2020tasks,fifty2021efficiently} aims to find groups of related tasks and employs multi-task learning to each group of tasks, with one model for each group. 
Our work focuses on training single-task models, but task grouping techniques can be used to look for possible friend tasks.

\noindent\textbf{Conversational Semantic Role Labeling} CSRL is a task for predicting the semantic roles of predicates in a conversational context. \citet{wu2021csagn} leveraged relational graph neural networks~\cite{schlichtkrull2018modeling} to model both the speaker and predicate dependency, achieving some promising results. However, the current dataset~\cite{xu2021conversational} for CSRL is limited to mono-domain. High-quality labeled data for new domains are needed to empower more applicable CSRL models.

\noindent\textbf{Dialogue Rewriting} DR is commonly framed as a sequence-to-sequence problem which suffers large search space issue~\cite{elgohary2019can,huang2021sarg}. To address it, \citet{hao2021rast} cast DR to sequence labeling, transforming rewriting an utterance as deleting tokens from an utterance or inserting spans from the dialogue history into an utterance. \citet{jin2022hierarchical} improved the continuous span issue in \cite{hao2021rast} by first generating multiple spans for each token and slotted rules and then replacing a fixed number rules with spans.

%% file: framework.tex
\section{Friend-training}
Friend-training is an iterative training framework to jointly refine models of several friend tasks. Different from self-training, friend-training injects cross-task supervision into the selection of pseudo-labels. We first briefly describe self-training before presenting friend-training.

\subsection{Self-training}
Classic self-training aims at iteratively refining a model of a single task by using both labeled data and a large amount of unlabeled corpus. 
At each iteration, the model first assigns the unlabeled data with pseudo-labels. Subsequently, a set of the unlabeled instances with pseudo-labels is selected for training, presumably with information for better model generalization. Then cross-entropy of model predictions and labels on both gold and pseudo-labeled data is minimized to update the model:
\begin{equation}\label{eq:loss}
    L = \sum_{i=1}^{N}y_{i}\log\frac{y_{i}}{p_{i}} + \lambda\sum_{i=1}^{N^{\prime}}y_{i}^{\prime}\log\frac{y_{i}^{\prime}}{p_{i}^{\prime}},
\end{equation}
where the left term is the loss for the labeled data and the right for the unlabeled data while $\lambda$ is a coefficient to balancing them; $N$($N^{\prime}$) is the number of instances, $y$ ($y^{\prime}$) is the label and $p$ ($p^{\prime}$) is the output probability of the model.

Self-training is usually limited to only one task, but there are thousands of NLP tasks already proposed and many of them are related. Models trained for one task can be great teachers for other related tasks. We explore this cross-task supervision in self-training by incorporating two novel modules introduced in subsection~\ref{sec:friend-training}.

\subsection{Friend-training}\label{sec:friend-training}

For friend-training with two tasks,\footnote{We focus on the two-friend version of friend-training in this work, however, friend-training can easily be extended to more than two friends.} we have two classifiers $f_a$ and $f_b$ trained on two different tasks with labeled training sets $\mathcal{L}_a$ and $\mathcal{L}_b$, with expected accuracies $\eta_a$ and $\eta_b$, respectively. The two datasets are created independently and the prediction targets of the two tasks are partially related through a pair of translation functions $\mathcal{F}_a: \hat{Y}_a \rightarrow \Sigma$ and $\mathcal{F}_b: \hat{Y}_b \rightarrow \Sigma$, where $\Sigma$ is the set of possible sub-predictions that all possible predictions of the two tasks $\hat{Y}_a$ and $\hat{Y}_b$ can be reduced to. $|\hat{Y}_a| \ge |\Sigma|, |\hat{Y}_b| \ge |\Sigma|$. We assume that the translation functions are general functions with the expected probability of generating a translation $\epsilon_{\mathcal{F}} = \frac{1}{|\Sigma|}$. The translation functions are deterministic and always map the gold labels of the friend tasks for the same input to the same translation.

Both classifiers make predictions on the unlabeled set $\mathcal{U}$ at iteration $k$. Some instances $\mathcal{U}_\mathcal{F}^k$ with pseudo-labels are chosen as new training data based on the results of the translation functions, $\phi_a(x) =\mathcal{F}_a(f_a(x))$ and $\phi_b(x) =\mathcal{F}_b(f_b(x))$, and some selection criteria, such as total agreement. If total agreement is used as the selection criterion, the probability of erroneous predictions for $f_a$ in these instances is

\begin{align}
    &\mathrm{Pr}_{x} [f_a(x) \ne f^*_a(x)| \mathcal{\phi}_a(x) = \mathcal{\phi}_b(x)] \nonumber\\
     =& 1- \frac{\eta_a \mathrm{Pr}_{x} [ \mathcal{\phi}_a(x) = \mathcal{\phi}_b(x) | f_a(x) = f^*_a(x)] }{\mathrm{Pr}_{x} [\mathcal{\phi}_a(x) = \mathcal{\phi}_b(x)]},
\label{eq:error_rate}
\end{align}
with $f^*$ being the optimal classifier.

Because both classifiers are very different due to training data, annotation guidelines, models, prediction targets, etc.,  being all different, the two classifiers are very likely to be independent of each other. Under this condition Equation \ref{eq:error_rate} becomes
\begin{align}\label{eq:match_prob}
    & 1- \frac{\eta_a (\eta_b + \epsilon_{\mathcal{F}} (1 - \eta_b) ) }{\mathrm{Pr}_{x} [\mathcal{\phi}_a(x) = \mathcal{\phi}_b(x)]} \nonumber \\
     = & 1- \frac{Z}{Z + \eta_b\epsilon_{\mathcal{F}} (1 - \eta_a) + E},
\end{align}
where $Z=\eta_a (\eta_b + \epsilon_{\mathcal{F}} (1 - \eta_b) )$ and $E = \epsilon_\mathcal{F}^2(1-\eta_a)(1-\eta_b)$. We give the detailed derivation of Equation~\ref{eq:error_rate} and \ref{eq:match_prob} in Appendix~\ref{sec:apx_error_rate}. This indicates that the quality of the picked instances is negatively correlated with the number of false positive instances brought by the noisy translation $\eta_b\epsilon_{\mathcal{F}} (1 - \eta_a)$, and the number of matching negative instances $E$. 
When $\epsilon_\mathcal{F}$ is minimized by choosing translation functions with a sufficiently large co-domain $\Sigma$, the probability of error instances chosen when two classifiers agree approaches 0. This also indicates that even when $1-\eta_a$ is large, i.e. $f_a$ performs badly, if the co-domain is large, the error rate of the chosen instances can still be kept very low.\footnote{Intuitively, this means independent classifiers trained to do different tasks are unlikely to predict the same but wrong sub-prediction for a given input, if the sub-prediction includes a large number of decisions.}
As the dependence between the two classifiers grows in training, the probability of error instances also increases.
When they are completely dependent on each other, Equation \ref{eq:error_rate} becomes $1-\eta_a$, i.e. classic self-training.

Based on this formulation, two additional modules are needed: (1) a \textit{translation matcher} that maps predictions of two models trained on different tasks into the same space and computes a matching score; (2) an \textit{augmented (instance) selector} which selects instances with pseudo-labels for the classifiers considering both the matching score of the translated predictions and the model confidences. 

\noindent\textbf{Translation Matcher}\label{sec:matcher}
Given the prediction of models of two friend tasks $f_a(x)$ and $f_b(x)$, the translation matcher $\mathcal{M}$ leverages translation functions $\mathcal{F}_a$ and $\mathcal{F}_b$ to get the translated pseudo-labels and computes a matching score $m$ for the pair of pseudo-labels, which represents the similarity of the pair in the translation space:
\begin{equation}\label{eq:matcher}
    m_{a,b}=\mathcal{M}\left(\mathcal{F}_a(f_a(x)),\mathcal{F}_b(f_b(x))\right),
\end{equation}
with total agreement being 1. This matching score serves as a criterion for the selection of high quality pseudo-labels with cross-task supervision. 

\noindent\textbf{Augmented Selector}\label{sec:selector}
Apart from pseudo-label similarity, other information about pseudo-label quality can be found from model confidence, which self-training algorithms specifically utilize, to augment matching scores. The augmented selector considers both the confidence of the pseudo-labels from task models, denoted as $\left\{c_a,c_b \right\}$, and the matching scores:
\begin{equation}\label{eq:selector}
    q_\tau = \mathcal{S}_\tau(m_{a,b}, c_\tau),
\end{equation}
where $q_\tau \in \left\{0, 1\right\}$ represents the selection result of the pseudo-label for task $\tau \in {a,b}$. Therefore, instances with low matching scores but high confidence may also be selected as the training data. 
The complete algorithm is shown in Algorithm \ref{algo:framework}.

\begin{algorithm}[t]
    \SetKwInOut{Input}{Input}\SetKwInOut{Output}{Output}
    \Input{Labeled data sets for two friend tasks, $\mathcal{L}_a,\mathcal{L}_b$; an unlabeled data set $\mathcal{U}$; task models $f_a,f_b$.}
    \Output{Refined $f_a,f_b$.}
    
    Pre-train $f_\tau$ with $\mathcal{L}_\tau$ $\left(\tau \in a,b\right)$;\\
    \While{not until the maximum iteration}{
        $\mathcal{L}^{u}_a=\emptyset$; $\mathcal{L}^{u}_b=\emptyset$; \\
        \For{$z$ in $\mathcal{U}$} {
            Generate $f_a(z),f_b(z)$ and $c_a,c_b$;\\
            $m_{a,b}$ $\leftarrow$ Equation~\ref{eq:matcher};\\
            $q_a,q_b$ $\leftarrow$ Equation~\ref{eq:selector};\\
            \uIf{$q_\tau=1$ $\left(\tau \in a,b\right)$} {
                $\mathcal{L}^{u}_\tau = \mathcal{L}^{u}_\tau + \left\{z, v_\tau\right\}$;
            }
        }
    Update $f_\tau$ with $\mathcal{L}_\tau,\mathcal{L}_{\tau}^u$ by Equation \ref{eq:loss} $\left(\tau \in a,b\right)$;
    }
    Return $f_a,f_b$;
    \caption{Two-task friend-training}\label{algo:framework}
\end{algorithm}

\section{Friend Training between CSRL and DR}
\label{csrl-rewriting}

\begin{figure*}[!t]
\centering
\includegraphics[width=\textwidth]{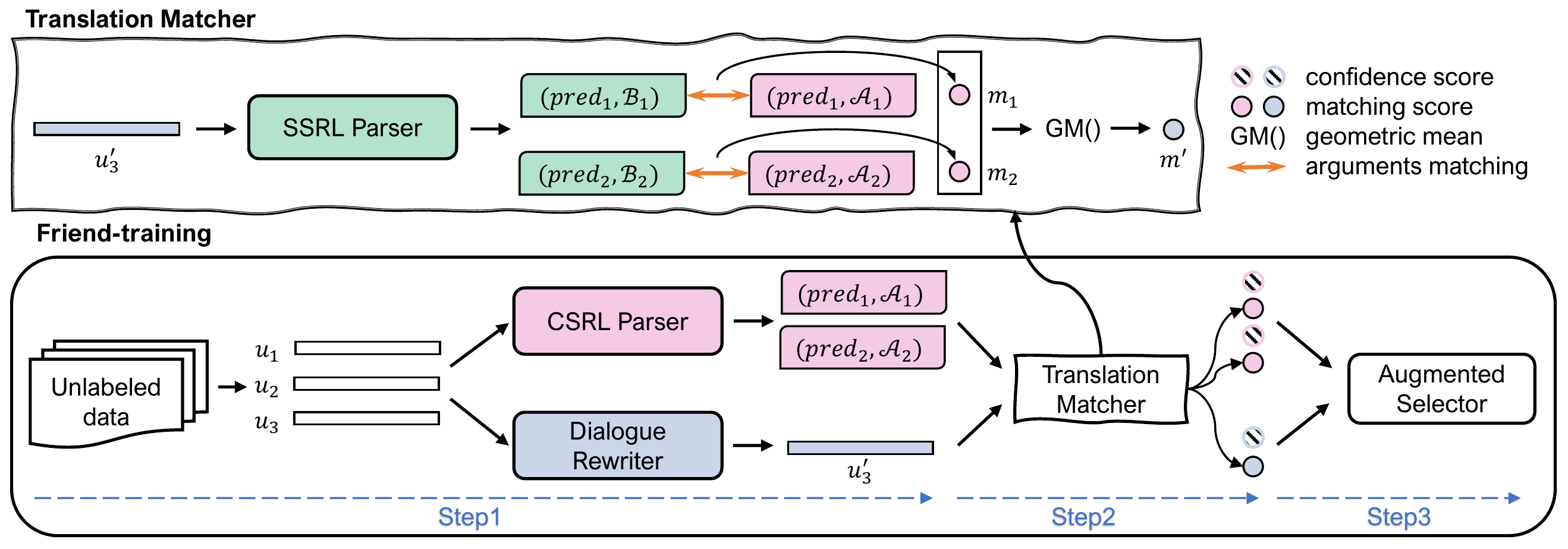}
\caption{The overview of the friend-training process between CSRL and DR for one dialogue instance which has three utterances and the last utterance contains two predicates. Step1: the unlabeled dialogue is labeled by the CSRL parser and dialogue rewriter, resulting in predictions of arguments for the predicates (CSRL) and the rewritten utterance (DR), respectively. Step2: Pseudo-labels of both tasks are fed into the translation matcher to get their matching scores: the translation matcher first conducts sentence-level semantic role labeling (SSRL) on the rewritten utterance $u_3^{\prime}$ and then compares the results with those of the CSRL parser for matching scores. Step3: The threshold-based augmented selector makes the final decision of whether to add each pseudo-label to the training data considering both their confidence and matching scores. Best viewed in color.}
\label{fig:overview}
\end{figure*}

To verify the effectiveness of friend-training, we select two dialogue understanding tasks as friend tasks to conduct friend-training experiments for a case study: conversational semantic role labeling (CSRL) and dialogue rewriting (DR). While both require skills such as coreference and zero-pronoun resolution, the two tasks focus on different properties of the dialogue utterance: (1) CSRL focuses on extracting arguments of the predicates in the utterance from the whole dialogue history; (2) DR aims to rewrite the last turn of a dialogue to make it context-free and fluent by recovering all the ellipsis and coreference in the utterance. Figure~\ref{fig:overview} provides an overview of friend-training between the above two tasks. Next, we first introduce the task models and then specify the translation matcher and augmented selector for applying friend-training.

\subsection{Task Models}
\noindent\textbf{Task Definition} A dialogue consists of $N$ temporally ordered utterances $\left\{u_1,...,u_N\right\}$. (1) Given utterance $u_t$ and $K$ predicates $\left\{\text{pred}_1,..., \text{pred}_K\right\}$ of $u_t$, a CSRL parser predicts spans from the dialogue as arguments for all predicates. (2) A dialogue rewriter rewrites $u_t$ to make it context-free according to its context $\left\{u_1,...,u_{t-1}\right\}$.\\
\noindent\textbf{Dialogue Encoder} We concatenate dialogue context $\left\{u_1,...,u_{t-1}\right\}$ and the current utterance $u_t$ as a sequence of tokens $\left\{x_1,...,x_M\right\}$ and encode it with BERT~\cite{devlin2018bert} to get the contextualized embeddings:
\begin{equation}
   \mathbf{E} = \mathbf{e}_1, ..., \mathbf{e}_M=\textrm{BERT}(x_1,...,x_M) \in \mathbb{R}^{H \times M}. \nonumber
\end{equation}
Encoders for CSRL and DR share no parameters, but for simplicity, we use the same notation $\mathbf{E}$ for their outputs.\\
\noindent\textbf{Conversational Semantic Role Labeling} With the contextualized embeddings, we further generate predicate-aware utterance representations $\mathbf{G} = \{\mathbf{g}_1, ..., \mathbf{g}_M\} \in \mathbb{R}^{H \times M}$ as \citet{wu2021csagn} by applying self-attention \cite{vaswani2017attention} to $\mathbf{E}$ with predicate-aware masking, where a token is only allowed to attend to tokens in the same utterance and tokens from the utterance containing the predicate:
\begin{equation}
\textrm{Mask}_{i,j} = 
    \begin{cases}
        1 & \text{if} \;u_{[i]} = u_{[j]} \; or \; u_{[j]} = u_{[pred]}, \\
        0 & \text{otherwise}, \nonumber
    \end{cases}
\end{equation}
where $u_{[m]}$ denotes the utterance containing token $x_m$ and $u_{[pred]}$ denotes the one with the predicate.

The predicate-aware representations are then projected by a feed-forward network to get the distribution of labels for each token:
\begin{equation}
    \mathbf{P}^c=\textrm{softmax}_{\textrm{column-wise}}(\mathbf{W}_c\mathbf{G} + \mathbf{b}_c) \in \mathbb{R}^{C \times M}, \nonumber
\end{equation}
where $\mathbf{W}_c$ and $\mathbf{b}_c$ are learnable parameters and $C$ is the number of labels. The labels follow BIO sequence labeling scheme: B-X and I-X respectively denote the token is the first token and the inner token of argument X, where O means the token does not belong to any argument. The output of the CSRL parser for $K$ predicates are denoted as $\left\{\mathcal{A}_1, ..., \mathcal{A}_K\right\}$, where set $\mathcal{A}_k$ containing the arguments for $\text{pred}_k$.\\
\noindent\textbf{Dialogue Rewriting} Following \citet{hao2021rast}, we cast DR as sequence labeling. Specifically, a binary classifier on the top of $\mathbf{E}$ first determines whether to keep each token for in utterance $u_t$ in the rewritten utterance:
\begin{equation}
    \mathbf{P}^d=\textrm{softmax}_{\textrm{column-wise}}(\mathbf{W}_d\mathbf{E} + \mathbf{b}_d) \in \mathbb{R}^{2 \times M}, \nonumber
\end{equation}
where $\mathbf{W}_d$ and $\mathbf{b}_d$ are learnable parameters. Next, a span of the context tokens is predicted to be inserted in front of each token. In practice, two self-attention layer~\cite{vaswani2017attention} are adopted to calculate the probability of context tokens being the start index or end index of the span:
\begin{align}
    \mathbf{P}^{st} &= \textrm{softmax}_{\textrm{column-wise}}(\textrm{Attn}_{st}(\mathbf{E})) \in \mathbb{R}^{M \times M}, \nonumber \\
    \mathbf{P}^{ed} &= \textrm{softmax}_{\textrm{column-wise}}(\textrm{Attn}_{ed}(\mathbf{E}))\in \mathbb{R}^{M \times M}, \nonumber
\end{align}
where $\mathbf{P}^{st}_{i,j}$ ($\mathbf{P}^{ed}_{i,j}$) denotes the probability of $x_i$ being the start (end) index of the span for $x_j$. Then by applying argmax to $\mathbf{P}$, we could obtain the start and end indexes of the span for each token:
\begin{align}
    \mathbf{s}^{st} = \textrm{argmax}_{\textrm{column-wise}}(\mathbf{P}^{st}) \in \mathbb{R}^{M}, \nonumber \\
    \mathbf{s}^{ed} = \textrm{argmax}_{\textrm{column-wise}}(\mathbf{P}^{ed}) \nonumber \in \mathbb{R}^{M},
\end{align}
The probability of the span to be inserted in front of $x_m$ is $\mathbf{P}^{st}_{\mathbf{s}_m^{st},m} \times \mathbf{P}^{ed}_{\mathbf{s}_m^{ed},m}$ when $\mathbf{s}_m^{st} \leqslant \mathbf{s}_m^{ed}$. When $\mathbf{s}_m^{st} > \mathbf{s}_m^{ed}$, it means no insertion. The output of the dialogue rewriter for $u_t$ is denoted as $u_t^{\prime}$.

\subsection{Translation Matcher}
To translate the outputs (pseudo-labels) from the CSRL parser $\left\{\mathcal{A}_1, ..., \mathcal{A}_K\right\}$ and the dialogue rewriter $u_t^{\prime}$ into a same space, we leverage a normal sentence-level semantic role parser with \textit{fixed} parameters to greedily extract arguments from the rewritten utterance $u_t^{\prime}$ for the $K$ predicates, denoted as $\left\{\mathcal{B}_1, ..., \mathcal{B}_K\right\}$ (Appendix~\ref{sec:apx_case} shows an example). So the common target space $\Sigma$ is the label space of CSRL, which is large enough to make the error rate of chosen instances keep very low (see the analysis in subsection~\ref{sec:friend-training}). The matching score $m_{k} \in [0,1]$ for $\text{pred}_k$ is calculated based on the edit distance between $\mathcal{A}_k$ and $\mathcal{B}_k$:
\begin{equation}
    m_{k} = 1 - \frac{\textrm{dist}(\oplus \mathcal{A}_k, \oplus \mathcal{B}_k)}{\textrm{max}(\textrm{len}(\oplus \mathcal{A}_k), \textrm{len}(\oplus \mathcal{B}_k))}, \nonumber
\end{equation}
where dist() calculates the edit distance between two strings, \textrm{len()} returns the length of a string and $\oplus \mathcal{A}_k$ denotes the concatenation of arguments in set $\mathcal{A}_k$ in a predefined order of arguments\footnote{Argument concatenating order: \texttt{ARG0}, \texttt{ARG1}, \texttt{ARG2}, \texttt{ARG3}, \texttt{ARG4}, \texttt{ARGM-TMP}, \texttt{ARGM-LOC}, \texttt{ARGM-PRP}} (empty strings means arguments do not exist). Furthermore, we obtain the overall matching score $m^{\prime} \in [0,1]$ for the rewritten utterance $u_t^{\prime}$ as follows:
\begin{equation}
    m^{\prime} = \textrm{GM}(m_1,...,m_K), \nonumber
\end{equation}
where GM() represents the geometric mean.

\subsection{Augmented Selector}
The augmented selector selects high-quality pseudo-labels according to both their matching scores and confidence. For CSRL, we calculate the confidence score for each predicate based on the output of the softmax layer. Specifically, we obtain the confidence of an argument for $\text{pred}_k$ by multiplying the probability of its tokens, denoted as $\{a_{k1}, ..., a_{k|\mathcal{A}_k|}\}$. We then use the geometric mean of all the confidence of arguments belonging to $\text{pred}_k$ as the confidence for $\text{pred}_k$. The overall score $s_{k} \in [0,1]$ for $\text{pred}_k$ is calculated as follows:
\begin{equation}
    s_k = \alpha \textrm{GM}(a_{k1}, ..., a_{k|\mathcal{A}_k|}) + (1 - \alpha) m_{k},\nonumber
\end{equation}
where hyper-parameter $\alpha$ gives a balance between the matching score and the confidence. For DR, we multiply the probabilities of spans to be inserted and of decisions on whether to keep tokens or not as the model confidence of $u_t^{\prime}$, denoted as $b_{t}$. The overall score $r_t \in [0,1]$ of $u_t^{\prime}$ is as follows:
\begin{equation}
    r_t = \beta b_t + (1 - \beta)m^{\prime},\nonumber
\end{equation}
where a larger value of hyper-parameter $\beta$ places more importance on the model confidence. $\alpha$ and $\beta$ are set to be 0.2 for both tasks in the experiments.

Pick thresholds are set for $s_k$ and $r_t$ to control the number and quality of selected pseudo-labels. We analyze the effects of different values of thresholds in subsection \ref{sec:abl}. 

%% file: experiment.tex
\section{Experiments}\label{sec:exp}
\begin{table*}[t!]
\begin{subtable}{1.0\textwidth}
    \centering
    \resizebox{0.9\textwidth}{!}{

    \begin{tabular}{lcccccc}
    \toprule
    
    {} & \multicolumn{3}{c}{WeiboCSRL} & \multicolumn{3}{c}{RESTORATION}  \\
    
    \cmidrule[0.5pt](rl){2-4}
    \cmidrule[0.5pt](rl){5-7}

    {\textbf{Method}} & {\textbf{Pre.}} & {\textbf{Rec.}} & {\textbf{F1}} & {\textbf{R-L}} & {\textbf{EM}} & {\textbf{WER}($\Downarrow$)} \\
    \midrule
    Base & 57.97 & 54.47 & 56.16 & 82.78 & 25.25 & 28.69  \\
    Multitask-Base & 53.66 & 54.32 & 53.99 & 81.68 &  22.49 & 32.44 \\
    SST~\cite{scudder1965probability} & 60.85 & 56.54 & 58.62 & 85.22 & 32.97 & 22.22  \\
    MT~\cite{tarvainen2017mean} & 58.42 & 55.71 & 57.03 & 83.76 & 28.82 & 26.49 \\
    CPS~\cite{chen2021semi} & 60.34 & 52.87 & 56.36 & 85.60 & 32.68 & 22.78 \\
    SCoT~\cite{blum1998combining} & 57.33 & 54.13 & 55.69  & 84.51 & 29.25 & 24.87 \\
    STBR~\cite{bhat2021self} & 60.77 & 58.04 & 59.38  & 85.79 & 33.78 & 23.30 \\
    STea~\cite{yu2021self} & 60.10 & 55.13 & 57.50  & 85.75 & 34.23 & 22.17  \\
    \midrule
    FDT (Ours) & \textbf{\makecell{65.29($\uparrow$4.44)}} & \textbf{\makecell{58.63($\uparrow$2.09)}} & \textbf{\makecell{61.78($\uparrow$3.16)}} & \textbf{\makecell{86.82($\uparrow$1.60)}} & \textbf{\makecell{38.22($\uparrow$5.25)}} & \textbf{\makecell{20.31($\uparrow$1.91)}} \\
    \bottomrule
    \end{tabular}
    }
    \caption{Domain generalization for models trained with DuConv and REWRITE.}
    \end{subtable}
\vspace*{0.3 cm}
\newline    %
    \begin{subtable}{1.0\textwidth}
    \centering
    \resizebox{0.9\textwidth}{!}{
    \begin{tabular}{lcccccc}
    \toprule
    
    
    {} & \multicolumn{3}{c}{DuConv} & \multicolumn{3}{c}{REWRITE} \\
    
    \cmidrule[0.5pt](rl){2-4}
    \cmidrule[0.5pt](rl){5-7}

    {\textbf{Method}} & {\textbf{Pre.}} & {\textbf{Rec.}} & {\textbf{F1}} & {\textbf{R-L}} & {\textbf{EM}} & {\textbf{WER}($\Downarrow$)} \\
    \midrule
    Base & 29.50 & 21.90 & 25.14 & 73.44 & 3.60 & 39.98 \\
    Multitask-Base & 22.43 & 20.63 & 21.49 & 78.97 & 11.70 & 40.46 \\
    SST~\cite{scudder1965probability}  & 34.16 & 27.49 & 30.46 & 80.93 & 27.80 & 31.02 \\
    MT~\cite{tarvainen2017mean} & 36.32 & 30.69 & 33.27 & 81.66 & 33.00 & 31.66 \\
    CPS~\cite{chen2021semi} & 37.14 & 29.47 & 32.86 & 79.56 & 23.30 & 32.60 \\
    SCoT~\cite{blum1998combining} & 38.37 & 26.15 & 31.10 & 78.58 & 22.31 & 33.79 \\
    STBR~\cite{bhat2021self} & 32.37 & 25.21 & 28.34 & 82.37 & 29.80 & 30.31 \\
    STea~\cite{yu2021self} & 39.34 & 28.78 & 33.25 & 83.04 & 31.57 & 30.36 \\
    \midrule
    FDT (Ours) & \textbf{\makecell{40.41($\uparrow$6.25)}} & \makecell{30.82($\uparrow$3.33)} & \makecell{34.97($\uparrow$4.51)} & \makecell{82.83($\uparrow$1.90)} & \makecell{34.20($\uparrow$6.40)} & \makecell{27.87($\uparrow$3.15)} \\
    FDT-S (Ours) & 40.12 & \textbf{33.41} & \textbf{36.46}  & \textbf{83.11} & \textbf{37.10} & \textbf{26.88} \\
    \midrule
    \textit{\textbf{Fully-trained} Base} & \textit{69.83} & \textit{68.53} & \textit{69.17} & \textit{89.47} & \textit{52.30} & \textit{20.54} \\
    \bottomrule
    \end{tabular}
    }
    \caption{Few-shot learning for models trained with DuConv and REWRITE.}
    \end{subtable}
\caption{Test results for domain generalization and few-shot learning. Base denotes the task models trained with data from a single task. Multitask-Base denotes the base model of CSRL and DR sharing the same dialogue encoder. Results are averaged across three runs. $\Downarrow$ means lower is better. For few-shot learning, performance of the base models trained with the full training set from the single task is provided for reference.}
\label{tab-main}
\end{table*}

\subsection{Setup}
\noindent\textbf{Datasets} We use five dialogue datasets in our experiments with domains spanning movies, celebrities, book reviews, products, and social networks. For CSRL, we use DuConv~\cite{xu2021conversational} and WeiboCSRL and for DR, REWRITE~\cite{su2019improving} and RESTORATION~\cite{pan2019improving}. The datasets of the same task differ in domains and sizes. WeiboCSRL is a newly annotated CSRL dataset for out-of-domain testing purposes. Moreover, we use LCCC-base~\cite{wang2020large} as the unlabeled corpus, which is a large-scale Chinese conversation dataset with 79M rigorously cleaned dialogues from various social media. More details on the annotation of WeiboCSRL and the properties of the datasets could be found in Appendix~\ref{sec:apx_dataset}.\\
\noindent\textbf{Experiment Scenarios} Our main experiments involve two scenarios. (1) Domain generalization: we use DuConv as the training data in the source domain and WeiboCSRL for out-of-domain evaluation, while for DR, REWRITE is used for training and RESTORATION for evaluation. 
(2) Few-shot learning: we randomly select 100 cases from DuConv and REWRITE as the training data for CSRL and DR, respectively, and conduct in-domain evaluation, which means models of both the tasks are co-trained with only a few samples of each task. The unlabeled data for both scenarios are 20k dialogues extracted from LCCC-base. Implementation details are provided in Appendix~\ref{sec:apx_detail}.

\noindent\textbf{Evaluation} We follow \citet{wu2021csagn} to report precision (Pre.), recall (Rec.), and F1 of the arguments for CSRL and \citet{hao2021rast} to report word error rate (WER)~\cite{morris2004and}, Rouge-L (R-L)~\cite{lin2004rouge} and the percent of sentence-level exact match (EM) for DR. 
\subsection{Baselines}\label{sec:baseline}

We compare friend-training with six semi-supervised training paradigms: two standard techniques such as standard self-training (SST)~\cite{scudder1965probability} and standard co-training (SCoT)~\cite{blum1998combining}, as well as four recent methods such as mean teacher (MT)~\cite{tarvainen2017mean}, cross pseudo supervision (CPS)~\cite{chen2021semi}, self-training with batch reweighting (STBR)~\cite{bhat2021self} and self-teaching (STea)~\cite{yu2021self}. See Appendix~\ref{sec:apx_baseline} for more details.

\subsection{Main Results}
Table~\ref{tab-main} shows the comparison between friend-training (FDT) and the baselines mentioned in subsection~\ref{sec:baseline}. FDT achieves the best overall performance over the baselines by significant margins in both domain generalization and few-shot learning scenarios, which demonstrates the effectiveness of FDT in different experimental situations to utilize large unlabeled corpora. Moreover, we show the absolute improvements of FDT over SST in parentheses $(\uparrow)$. As we could see, in few-shot learning, FDT obtain 4.51 and 3.15 higher absolute points over SST on F1 of DuConv and WER of REWRITE, respectively, than those of domain generalization, which are 3.16 and 1.91 points, revealing that FDT could realize its potential easier in few-shot learning. Besides, for few-shot learning, we further consider the situation where a full-trained base model from the friend task is available, denoted as FDT-S. As we could see, when the target task is CSRL, FDT-S makes a gain of 1.49 points on F1 over FDT and when the target task is DR, FDT-S outperforms FDT on WER by 0.99 points and EM by 2.90 points, indicating that more reliable supervision from friend task could further enhance the few-shot learning of the target task.

\begin{figure*}[!htbp]
\centering
    \begin{subfigure}[t]{0.31\textwidth}
    \centering
    \includegraphics[width=\textwidth]{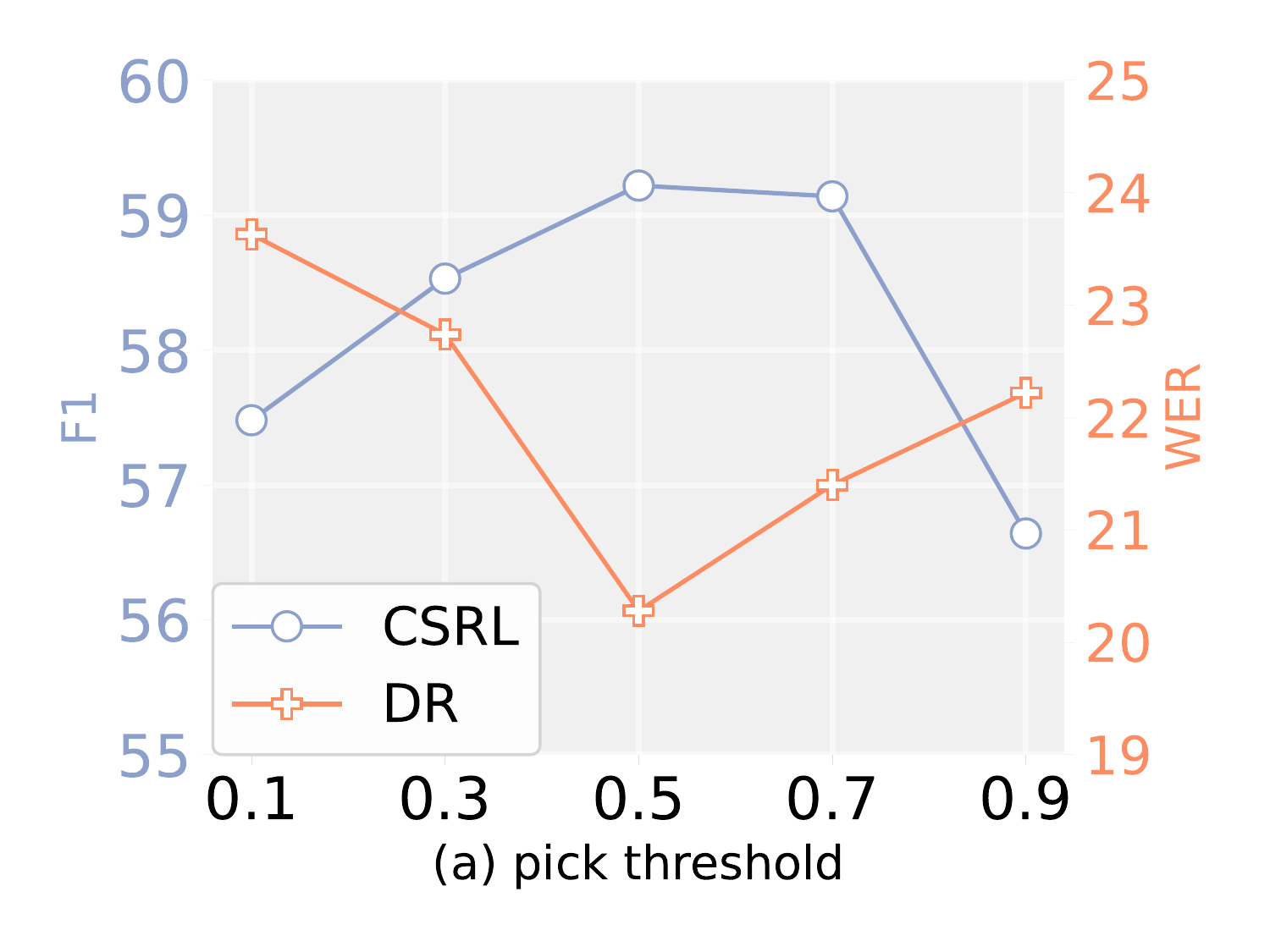}
    \caption{The effect of pick thresholds.}\label{fig:thres}
    \end{subfigure}
    \quad
    \begin{subfigure}[t]{0.31\textwidth}
    \centering
    \includegraphics[width=\textwidth]{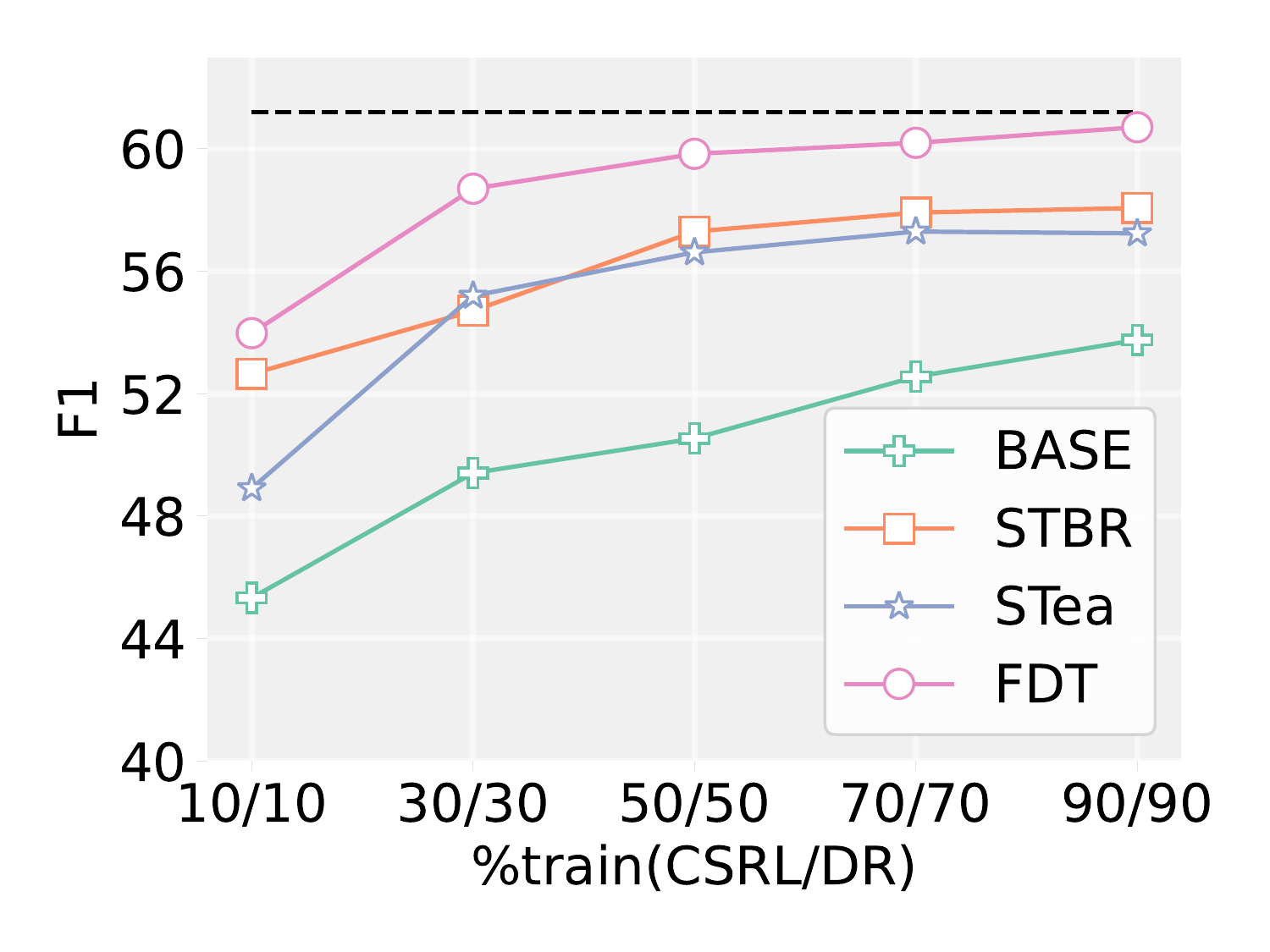}
    \caption{F1 on CSRL test set.}\label{fig:tea_csrl}
    \end{subfigure}
    \quad
    \begin{subfigure}[t]{0.31\textwidth}
    \centering
    \includegraphics[width=\textwidth]{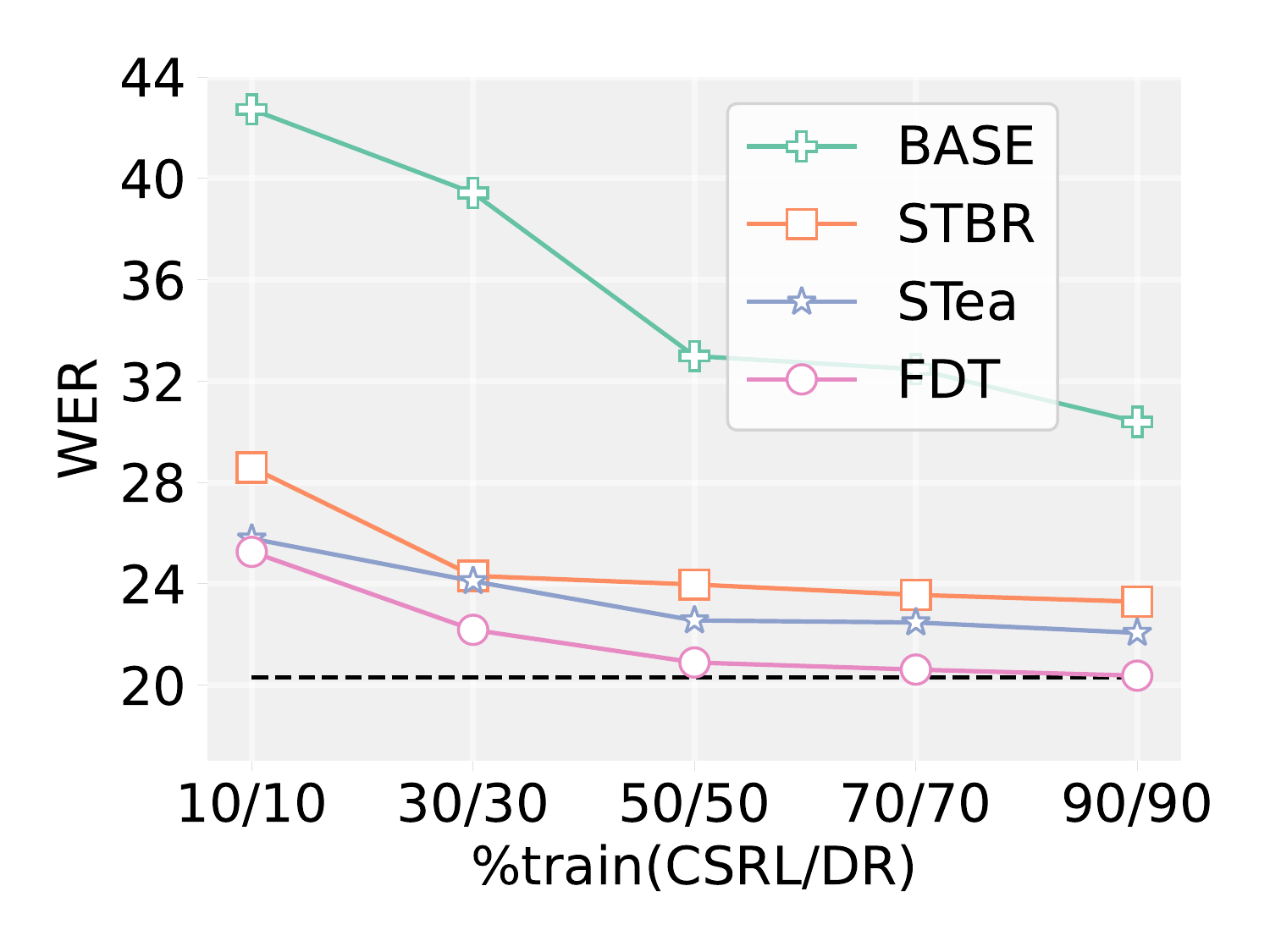}
    \caption{WER on DR test set.}\label{fig:tea_rewr}
    \end{subfigure}

\caption{Sub-figures (b) and (c) show the model performance of the comparing methods with different strengths of base models; the dashed horizontal line represents the performance of FDT with a fully trained base model.}
\end{figure*}

\begin{figure}[!htbp]
\centering
    \begin{subfigure}[b]{\columnwidth}
    \centering
    \includegraphics[width=\textwidth]{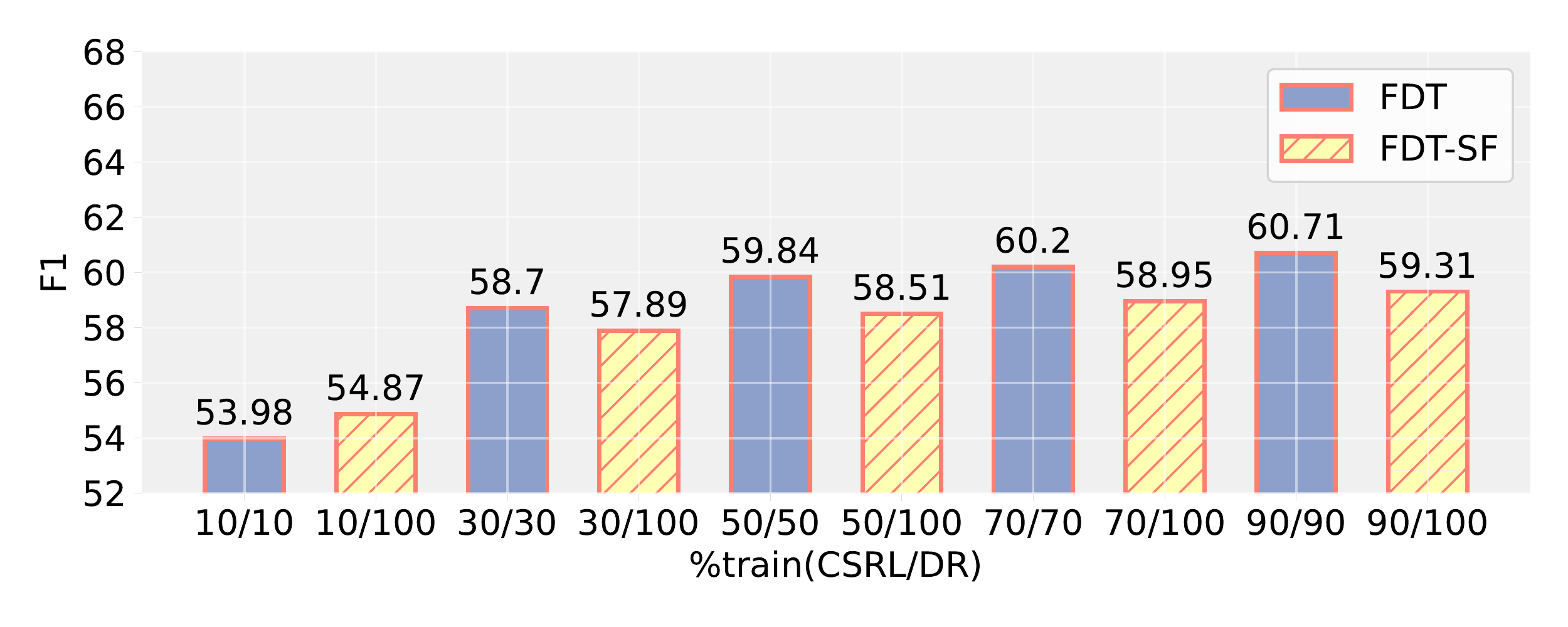}
    \caption{The performacne on CSRL test set.}
    \end{subfigure}
    \quad
    \begin{subfigure}[b]{\columnwidth}
    \centering
    \includegraphics[width=\textwidth]{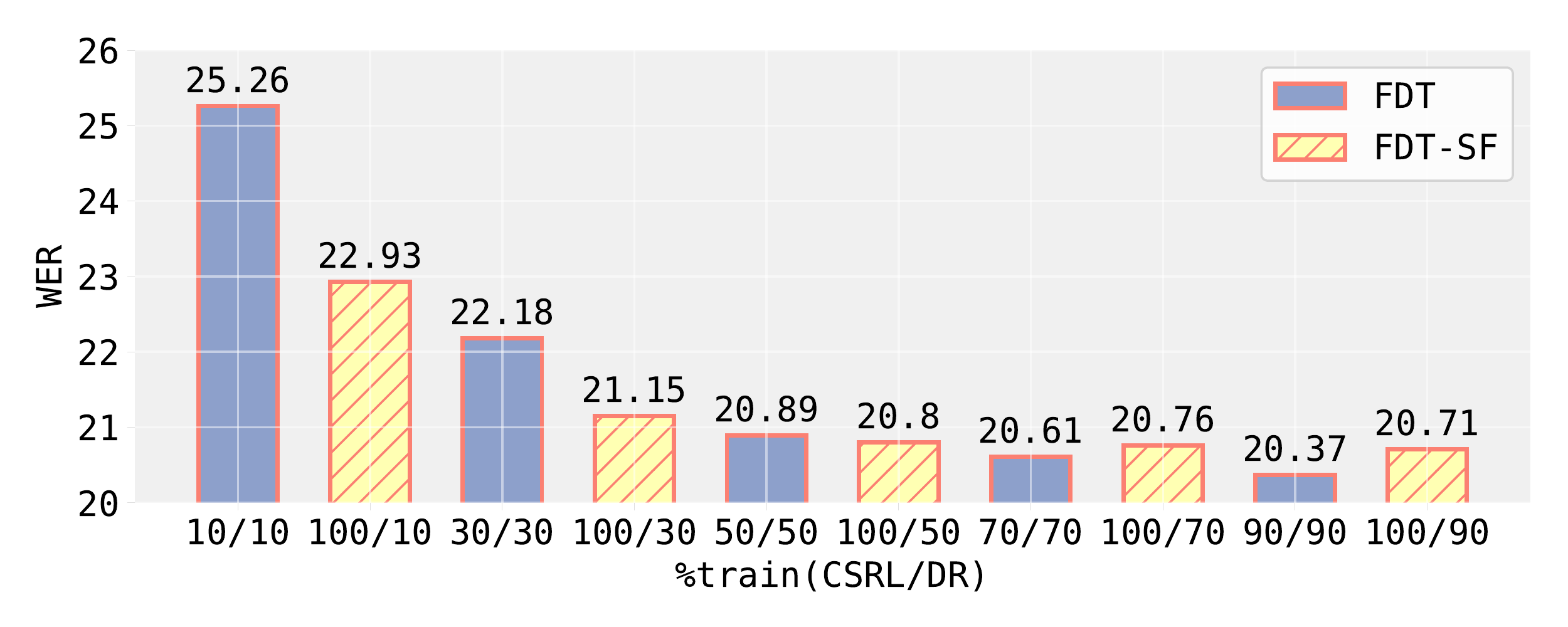}
    \caption{The performance on DR test set.}
    \end{subfigure}
\caption{The role of co-updating in friend-training.}\label{fig:fixabl}
\end{figure}

\subsection{Analysis}\label{sec:abl}
In this section, we conduct experiments to analyze how selected parameters and settings interact with model performance in FDT. \\
\noindent\textbf{Pick Thresholds} We vary the pick thresholds of CSRL and DR in domain generalization scenario and track the model performance: we fix the pick threshold of the friend task to the best (see Appendix~\ref{sec:apx_detail}) when varying that of the evaluating task. As illustrated in Figure \ref{fig:thres}, when the thresholds increase gradually, the models become better with higher F1 for CSRL and lower WER for DR. We attribute this to wrong pseudo-labels being filtered out by the augmented selector of FDT. Then the model performances hit the peaks and drop as the thresholds keep increasing in the interval of high values, which is owed to high thresholds producing insufficient pseudo-labels for iterative training. Automatically choosing proper pick thresholds is worth to be explored in the future. 

\noindent\textbf{The Strength of Base Model}
To understand and compare how performance of models before friend-training or self-training influences their final performance, we compare STBR, STea and FDT with the base models trained on different percentages of labeled data in the source domain when evaluating on out-domain testing data. Specifically, we follow domain generalization settings and use a variable percentage of labeled data to conduct experiments.

For CSRL and DR, respectively, we set the amount of labeled data as \{10\%/10\%, 30\%/30\%, 50\%/50\%, 70\%/70\%, 90\%/90\%\}.
The results are shown in Figure~\ref{fig:tea_csrl} and Figure~\ref{fig:tea_rewr}. We can see that all the methods adopting self-training to make use of unlabeled data surpass the base model by a significant margin, whether when given a weak or strong base model, demonstrating the effectiveness of self-training paradigm. Moreover, FDT achieves the best results across the evaluating percentages of labeled data: when the base model has a good amount of training data, such as those trained on 30\% labeled data and above, the performance of FDT is significantly better than STBR and STea, proving that FDT leverages the features learned from labeled data more effectively with cross-task supervision.\\
\noindent\textbf{The Role of Co-updating} We also explore the case where one of the models of the friend tasks is fully trained and does not have to be updated. We consider FDT-SF, FDT with a \textit{fixed} fully trained base model from the friend task in domain generalization\footnote{Specifically, when the evaluating task is CSRL, the amount of labeled data for the two tasks are set as \{10\%/100\%, 30\%/100\%, 50\%/100\%, 70\%/100\%, 90\%/100\%\}, and when the evaluating task is DR, \{100\%/10\%, 100\%/30\%, 100\%/50\%, 100\%/70\%, 100\%/90\%\}.}. As illustrated in Figure~\ref{fig:fixabl}, FDT-SF surpasses FDT when given a weak base model for the evaluating task because of the strong supervision from the friend task. However, FDT outperforms FDT-SF when the evaluating task is given a fairly-trained model, which demonstrates the benefits of co-updating the models in friend-training.  

%% file: conclusion.tex
\section{Conclusion}
We propose friend-training, the first cross-task self-training framework, which leverages supervision from friend tasks for better selection of pseudo-labels. Moreover, we provide specific modeling of friend-training between conversational semantic role labeling and dialogue rewriting. Experiments on domain generalization and few-shot learning scenarios demonstrate the promise of friend-training, which outperforms prior classical or state-of-the-art semi-supervised methods by 
substantial margins. 

%% file: limitation.tex
\section{Limitation}
We showed how the friend-training strategy can be applied to two dialogue understanding tasks in the case study here, but many other task pairs or task sets can be examined to fully explore the generality of the approach. Identifying friend tasks depends on expert knowledge in this work, but approaches for task grouping and task similarity may be used to automatically discover friend tasks. Besides, with the proliferation of cross-modal techniques, tasks of different modalities are expected to act as friend tasks as well. Also, designing translation functions and matchers for friend tasks in the friend-training framework requires an understanding of the relationship between the friend tasks, but prompting and model interpretability methods could potentially be applied for easing this process.

\section{Acknowledgement}
We thank the anonymous reviewers for their helpful comments and the support of National Nature Science Foundation of China (No.62176174).

%% file: appendix.tex
\clearpage 
\section{Appendix}

\subsection{Error rates}\label{sec:apx_error_rate}
We have two classifiers $f_a$ and $f_b$ trained on two different tasks with labeled training sets $\mathcal{L}_a$ and $\mathcal{L}_b$, with expected accuracies $\eta_a$ and $\eta_b$, respectively. The prediction targets of the two tasks are partially related through a pair of translation functions $\mathcal{F}_a: \hat{Y}_a \rightarrow \Sigma$ and $\mathcal{F}_b: \hat{Y}_b \rightarrow \Sigma$, where $\Sigma$ is the set of possible sub-predictions that all possible predictions of the two tasks $\hat{Y}_a$ and $\hat{Y}_b$ can be reduced to. $|\hat{Y}_a| \ge |\Sigma|, |\hat{Y}_b| \ge |\Sigma|$. The sub-predictions can be a part of the whole prediction targets for both tasks, or some lossy transformation of the prediction targets. For example, a sub-prediction for a POS-tagging task can be the POS tag of the first word (a part of the prediction) or the number of the NN tag in the whole prediction sequence (a transformation of the prediction). We assume that the translation functions are general functions with the expected probability of generating a translation $\epsilon_{\mathcal{F}} = \frac{1}{|\Sigma|}$; they are deterministic and always map the gold labels of the friend tasks for the same input to the same translation.
Both classifiers make predictions on the unlabeled set $\mathcal{U}$ at iteration $k$. Some instances $\mathcal{U}_\mathcal{F}^k$ with pseudo-labels are chosen as new training data based on the results of the translation functions, $\phi_a(x) = \mathcal{F}_a(f_a(x))$ and $\phi_b(x) =\mathcal{F}_b(f_b(x))$, and some selection criteria, such as total agreement. If total agreement is used as the selection criterion, the probability of erroneous predictions for $f_a$ in these instances is
\begin{align}
    &\mathrm{Pr}_{x} [f_a(x) \ne f^*_a(x)| \phi_a(x) = \phi_b(x)] \nonumber\\
    =& 1- \mathrm{Pr}_{x} [ f_a(x) = f^*_a(x) | \phi_a(x) = \phi_b(x)] \nonumber\\
    =& 1- \mathrm{Pr}_{x} [ f_a(x) = f^*_a(x)] \cdot \nonumber\\ &\frac{\mathrm{Pr}_{x} [ \phi_a(x) = \phi_b(x) | f_a(x) = f^*_a(x)] }{\mathrm{Pr}_{x} [\phi_a(x) = \phi_b(x)]} \nonumber\\
     =& 1 - \eta_a \cdot \nonumber\\& \frac{\mathrm{Pr}_{x} [ \phi_a(x) = \phi_b(x) | f_a(x) = f^*_a(x)] }{\mathrm{Pr}_{x} [\phi_a(x) = \phi_b(x)]},
\label{eq:error_rate_appendix}
\end{align}
with $f^*$ being the optimal classifier.
If we consider the two classifiers very likely to be independent from each other, then the probability of the translation of the predictions from the two classifiers being the same given the prediction from classifier $f_a$ is correct, which is $\mathrm{Pr}_{x} [ \phi_a(x) = \phi_b(x) | f_a(x) = f^*_a(x)]$, is the sum of the probability of the classifier $f_b$ making the correct prediction $\eta_b$ and the probability of an erroneous translation of the wrong prediction $\epsilon_\mathcal{F}(1-\eta_b)$. The probability of the translations matching $\mathrm{Pr}_{x} [\phi_a(x) = \phi_b(x)]$ has four situations: both predictions of the two classifiers are correct $\eta_a \eta_b$; $f_a(x)$ is correct but $f_b(x)$ is wrong and being translated erroneously $\eta_a \epsilon_{\mathcal{F}} (1 - \eta_b)$; $f_b(x)$ is correct but $f_a(x)$ is wrong and being translated erroneously $\eta_b \epsilon_{\mathcal{F}} (1 - \eta_a)$; both $f_a(x)$ and $f_b(x)$ are wrong but matching in the translation space $\epsilon_\mathcal{F}^2(1-\eta_a)(1-\eta_b)$.
Under these conditions Equation \ref{eq:error_rate_appendix} becomes
\begin{align}
    & 1- \frac{\eta_a (\eta_b + \epsilon_{\mathcal{F}} (1 - \eta_b) ) }{\mathrm{Pr}_{x} [\phi_a(x) = \phi_b(x)]} \nonumber \\
     = & 1- \frac{Z}{Z + \eta_b\epsilon_{\mathcal{F}} (1 - \eta_a) + E},
\end{align}
where $Z=\eta_a (\eta_b + \epsilon_{\mathcal{F}} (1 - \eta_b) )$ and $E = \epsilon_\mathcal{F}^2(1-\eta_a)(1-\eta_b)$ which shows that the term $\eta_b\epsilon_{\mathcal{F}} (1 - \eta_a)+E$ needs to be small to make the probability of matching translations with predictions being wrong small. This indicates that the quality of the picked instances based on the total agreement criterion is negatively correlated with the number of false positive instances brought by the noisy translation $\eta_b\epsilon_{\mathcal{F}} (1 - \eta_a)$, and the number of matching negative instances $E$. 
$\epsilon_\mathcal{F}$ can be minimized by choosing translation functions with a sufficiently large co-domain $\Sigma$, which means that when the translation space is large enough, it is unlikely that the two classifiers totally agree in the translation space but do not agree in their own prediction target spaces. So the probability of them agreeing and making correct predictions is much larger than agreeing but making incorrect predictions while the probability of error instances chosen when two classifiers agree approaches 0, indicating that even when $1-\eta_a$ is large, i.e. $f_a$ performs badly, if the co-domain is large, the error rate of the chosen instances can still be kept very low.

\subsection{Datasets}\label{sec:apx_dataset}
\noindent\textbf{Annotation Procedure of WeiboCSRL} The dialogues we use for CSRL annotation are extracted from LCCC-base~\cite{wang2020large}, which consists of at least 4 turns and 80 total characters to assure enough context for CSRL and DR. These dialogues and those used as unlabeled data for experiments in section~\ref{sec:exp} are from different parts of LCCC-base. For each dialogue, we annotate the predicates in the last utterance with the guidance of frame files of Chinese Proposition Bank\footnote{\url{https://verbs.colorado.edu/chinese/cpb/}}. For each predicate, the arguments we annotate are numbered arguments \texttt{ARG0-ARG4} and adjuncts \texttt{ARGM-LOC}, \texttt{ARGM-MNR}, \texttt{ARGM-TMP} and \texttt{ARGM-NEG}, whose definitions are shown in \cite{xue2006semantic}. \texttt{ARGM-MNR} is not included for evaluation in section~\ref{sec:exp} because annotation of \texttt{ARGM-MNR} is lacking in DuConv, the training data for CSRL. In the end, we obtain 3891 annotated predicates. \\
\noindent\textbf{Dataset Details} Table~\ref{tab:apx_dataset} shows the statistic of the datasets used in the experiments. DuConv~\cite{xu2021conversational} focuses on movies and celebrities and we adopt the same train/dev/test splitting as \citet{xu2021conversational}. REWRITE~\cite{su2019improving} contains 20K dialogues with a wide range of topics crawled from Chinese social media platforms; the last utterance of each dialogue is rewritten to recover all co-referred and omitted information.
RESTORATION~\cite{pan2019improving} contains dialogues from Douban\footnote{\url{https://www.douban.com}}, most of which are book, movie or product reviews. Compared with REWRITE, it contains more annotated dialogues, but around 40\% of the last utterances require no rewriting. 
\begin{table}[!h]
    \centering
    \resizebox{\columnwidth}{!}{
    \begin{tabular}{llr}
        \toprule
         & \textbf{Domain} & \textbf{\#Instance}(train/dev/test) \\
         \midrule
        \textbf{DuConv} & movies and celebrities & 23361 / 2852 / 2977 \\
        \textbf{WeiboCSRL} & social media & - / 1945 / 1946 \\
        \textbf{REWRITE} & social media & 16925 / 1000 / 1000 \\
        \textbf{RESTORATION} & \makecell{book, movie and \\product reviews} & - / 5000 / 5000 \\
        \bottomrule
    \end{tabular}
    }
    \caption{Dataset statistics.}
    \label{tab:apx_dataset}
\end{table}

\subsection{Implementation Details}\label{sec:apx_detail}
Dataset configuration of the tasks for the experimental scenarios are shown in Table~\ref{tab:apx_scenario}. \\
\noindent\textbf{Preprocessing Details} The maximum length of the input dialogue is set to 125. We transform the word-based labeling of DuConv to character-based labeling and we use the scripts\footnote{\url{https://github.com/freesunshine0316/RaST-plus}} provided by \citet{hao2021rast} to generate token-level annotations for sequence-labeling-based DR. For unlabeled data, we discard dialogues with less than 4 turns to guarantee sufficient context for CSRL and DR. \\ 
\noindent\textbf{Model Details} We use pretrained BERT\footnote{\url{https://huggingface.co/bert-base-chinese}}~\cite{devlin2018bert} as the dialogue encoder for CSRL and DR. Both the values of hyper-parameter $\alpha$ and $\beta$ are set to 0.2 and the pick thresholds are set to 0.6. We choose a state-of-the-art sentence-level semantic role labeling (SSRL) parser\footnote{\url{https://github.com/hankcs/HanLP}} for the translation matcher which follows the same structure as \cite{he-choi-2021-stem}.\\
\noindent\textbf{Training Details} We adopt AdamW~\cite{loshchilov2017decoupled} to optimize models with a learning rate of 4e-5 and batch size of 16. We use $\lambda=1$ to balance the loss of labeled and unlabeled data.
\begin{table}[!h]
    \centering
    \resizebox{\columnwidth}{!}{
    \begin{tabular}{llll}
        \toprule
        & \textbf{Task} & \textbf{Train} & \textbf{Dev\&Test} \\
        \midrule
        \textbf{\multirow{2}{*}{\makecell{DG}}} & CSRL & DuConv (train) & WeiboCSRL (dev,test) \\
        \cmidrule(lr){2-4}
        & DR & REWRITE (train) & RESTORATION (dev,test) \\
        \midrule
        \textbf{\multirow{2}{*}{\textbf{FSL}}} & CSRL & DuConv (100 cases) & DuConv (dev,test) \\
        \cmidrule(lr){2-4}
        & DR & REWRITE (100 cases) & REWRITE (dev,test) \\
        \bottomrule
    \end{tabular}
    }
    \caption{Dataset configuration of domain generalization (DG) and few-shot learning (FSL).}
    \label{tab:apx_scenario}
\end{table}

\begin{table*}[t!]
\centering
\resizebox{\textwidth}{!} {
    \begin{tabular}{lll}
    \toprule
    Context: 
    &
    \multicolumn{2}{l}{\makecell[l]{ch: [A]我有一个非常喜欢的女明星。[B]她叫什么名 字？[A]布蕾克·莱弗利。[B]她很有名吗？\\ en: [A] I have a favorite actress. [B] What's her name? [A] Blake Lively. [B] Is she famous?}} \\
    \midrule
    Current utterance
    & 
    \multicolumn{2}{l}{\makecell[l]{ch: [A]她是一个非常受关注的女明星。\\en: [A] She is a actress attracting much attention.}} \\
    \midrule
    Rewritten utterance
    &
    \multicolumn{2}{l}{\makecell[l]{ch: [A] 布蕾克·莱弗利是一个非常受关注的女明星。\\ en: [A] Blake Lively is a actress attracting much attention.}} \\
    \midrule
    Predicates & 是 (is) & 受 (attract) \\
    \midrule
    CSRL
    & 
    \makecell[l]{ch: ARG1: 一个非常受关注的女明星 \\ en: ARG1: a actress attracting much attention} 
    &
    \makecell[l]{ARG0: 布蕾克·莱弗利, ARG1: 关注 \\ ARG0: Blake Lively, ARG1: attention} \\
    \midrule
    \makecell[l]{SSRL}
    &
    \makecell[l]{ch: ARG0: 布蕾克·莱弗利,  ARG1: 一个非常受关注的女明星\\ en: ARG0: Blake Lively,  ARG1: a actress attracting much attention}
    &
    \makecell[l]{ARG0: 布蕾克·莱弗利, ARG1: 关注 \\ ARG0: Blake Lively,  ARG1: attention}
    \\
    \midrule
    Predicate matching score & 0.61 & 1.0 \\
    \midrule
    Predicate confidence & 0.95 & 0.54 \\
    \midrule
    Predicate overall score & 0.67 & 0.90 \\
    \midrule
    Utterance matching score & \multicolumn{2}{l}{0.81} \\
    Utterance confidence & \multicolumn{2}{l}{0.92} \\
    Utterance overall score & \multicolumn{2}{l}{0.83} \\
    \bottomrule
    \end{tabular}
}
\caption{Case study: [A] and [B] are the signatures of speakers. ch and en are the language abbreviations.}
\label{tab:case}
\end{table*}

\subsection{Baselines}\label{sec:apx_baseline}
Standard self-training~\cite{scudder1965probability} generates pseudo-labels to unlabeled data with a base model and uses them to train a new base model, which is repeated until convergence. Standard co-training ~\cite{blum1998combining} is similar to Standard self-training, but with two different base models dealing with the same task, generating pseudo-labels and adding the trusted ones for iterative training. Mean teacher~\cite{tarvainen2017mean} maintains a teacher model on the fly, whose weights are the exponential moving average of the weights of a student model across iterations. Cross pseudo supervision ~\cite{chen2021semi}, a state-of-the-art variant of self-training, maintains two networks with different initialization; the pseudo-label of one network is used to supervise the other network. Self-training with batch reweighting~\cite{bhat2021self} is a state-of-the-art self-training method that reweights the pseudo-labels in a batch when training according to the confidence from the teacher model. Self-teaching~\cite{yu2021self}, a state-of-the-art semi-supervised method that sequentially trains a junior teacher, a senior teacher and an expert student to leverage the unlabeled data.

For the hyper-parameters of the baselines, we keep the common hyper-parameters, such as learning rate, batch size, optimizer, and so on, the same as our proposed method. And we adopt the values of method-specific hyper-parameters used in the original papers, such as the merging weight of soft and hard labels of self-teaching and the smoothing parameter for updating of mean teacher.

\subsection{Case Study}\label{sec:apx_case}
We show a representative case of selecting pseudo-labels in Table~\ref{tab:case}. There are two predicates in current utterance: 是(is) and 受(attract). For 是(is), the CSRL parser yields only ARG1 while SSRL parser gives the same ARG1 but more of ARG0 based on the rewritten utterance. With the difference in arguments, the overall score is not high and this predicate could be regarded as low-quality if a high pick threshold is set. For 受(attract), the CSRL and SSRL parsers give the same arguments, which are the right answer. However, if we only consider the model confidence of the predicate, which is 0.54, this high-quality predicate are more likely to been discarded than consider the overall score, which is 0.90. And the rewritten utterance gets a high overall score, which is what we expected.

\subsection{Discussion on Generalization of the Framework}\label{notes_generalization}
It is not uncommon at all for different language tasks sharing some information. With one case study presented in detail in the main body of the paper, we also provide a short example of a different friend task pair -- constituency parsing and dependency parsing -- and explain how they can help each other and show the general nature of the friend-training framework.

Early work \citep{Magerman1995-ga,Collins2003-yh} has shown relationship between dependency and constituency parsing through head-finding rules, and \citet{Jin2019-so} show directly how common structures between dependency and constituency trees can be derived for parsing evaluation. In a dependency graph, a set of nodes with a single incoming edge is usually indicative of a phrase structure, such as a noun phrase, a verb phrase or a prepositional phrase. Such phrasal structures are well-marked in constituency treebanks, and could be used as the shared friend information for friend-training. Here is a sketch of how friend-training can be applied to this pair:
\begin{enumerate}
    \item Train a constituency parser and a dependency parser, presumably trained with a small number of training instances, as the models for the friend-training framework.
    \item Run both parsers on a common set of unlabeled data for parsing results.
    \item Find phrases such as noun, verb or prepositional phrases in the predicted constituency trees.
    \item Compare with the dependency trees, and check if spans of such phrases have only a single incoming edge. If so, the constituency and dependency parsing results can be considered agreeing, and added to the silver training set. If not, the silver annotation is discarded.
    \item Train the parsers again with the gold and silver training instances.
\end{enumerate}

As long as some shared information can be identified between two seemingly different tasks, the noisy agreement between that partial target can provide valuable supervision between two tasks. The translation and matching between constituency-dependency targets are simpler compared to the CSRL-rewriting pair presented in the paper, partly because no model is required for the translation process. However the CSRL-rewriting pair is more significant because heuristics may be difficult or not obvious to design where `bridging' tasks such as single-sentence SRL may be readily available.